\documentclass[runningheads]{llncs}
\pdfoutput=1

\usepackage{graphicx}
\usepackage[misc]{ifsym}
\usepackage{amsmath,amssymb,amsfonts}
\usepackage{algorithmic}

\usepackage{array}
\usepackage{boldline,multirow}
\usepackage{ragged2e}

\usepackage[caption=false,font=normalsize,labelfont=sf,textfont=sf]{subfig}
\begin{document}
\title{Correlation-aware Deep Generative Model for Unsupervised Anomaly Detection}
\titlerunning{Correlation-aware DGM for Unsupervised Anomaly Detection}

\author{Haoyi Fan\inst{1} \and Fengbin Zhang\inst{1} \and Ruidong Wang\inst{1} \and Liang Xi\inst{1} \and Zuoyong Li\inst{2}}

% \author{Anonymous PAKDD Submission Paper ID: 335\\
    % Content Area: Anomaly detection and analytics $\rightarrow$ Deep learning theory and applications in KDD}

\authorrunning{Fan et al.}

\institute{School of Computer Science and Technology, Harbin University of Science and Technology, Harbin 150080, China. \\
\email{\{isfanhy, zhangfengbin\}@hrbust.edu.cn,1820400010@stu.hrbust.edu.cn, xiliang@hrbust.edu.cn}\\
\and Fujian Provincial Key Laboratory of Information Processing and Intelligent Control, Minjiang University, Fuzhou 350121, China. \\
\email{fzulzytdq@126.com} \\
Corresponding authors: Fengbin Zhang, Zuoyong Li.
}

\maketitle              % typeset the header of the contribution
\begin{abstract}
Unsupervised anomaly detection aims to identify anomalous samples from highly complex and unstructured data, which is pervasive in both fundamental research and industrial applications. However, most existing methods neglect the complex correlation among data samples, which is important for capturing normal patterns from which the abnormal ones deviate. In this paper, we propose a method of \textbf{C}orrelation aware unsupervised \textbf{A}nomaly detection via \textbf{D}eep \textbf{G}aussian \textbf{M}ixture \textbf{M}odel (\textbf{CADGMM}), which captures the complex correlation among data points for high-quality low-dimensional representation learning. More specifically, the relations among data samples are correlated firstly in forms of a graph structure, in which, the node denotes the sample and the edge denotes the correlation between two samples from the feature space. Then, a dual-encoder that consists of a graph encoder and a feature encoder, is employed to encode both the feature and correlation information of samples into the low-dimensional latent space jointly, followed by a decoder for data reconstruction. Finally, a separate estimation network as a Gaussian Mixture Model is utilized to estimate the density of the learned latent vector, and the anomalies can be detected by measuring the energy of the samples. Extensive experiments on real-world datasets demonstrate the effectiveness of the proposed method.

\keywords{Anomaly Detection \and Graph Attention \and Gaussian Mixture Model \and Data Correlation.}
\end{abstract}

\section{Introduction}
\label{sec:introduction}
Anomaly detection aims at identifying abnormal patterns that deviate significantly from the normal behavior, which is ubiquitous in a multitude of application domains, such as cyber-security \cite{tan2011fast}, medical care \cite{xu2018cxnet}, and surveillance video profiling \cite{sultani2018real}. Formally, anomaly detection problem can be viewed as density estimation from the data distribution \cite{zong2018deep}: anomalies tend to reside in the low probability density areas. Although anomaly detection has been well-studied in the machine learning community, how to conduct unsupervised anomaly detection from highly complex and unstructured data effectively, is still a challenge.

Unsupervised anomaly detection aims to detect outliers without labeled data for the scenario that only a small number of labeled anomalous data combined with plenty of unlabeled data are available, which is common in real-world applications. Existing methods for unsupervised anomaly detection can be divided into three categories: reconstruction based methods, clustering based methods, and one-class classification based methods. Reconstruction based methods, such as PCA \cite{jolliffe2003principal} based approaches \cite{xu2010robust,pascoal2012robust} and autoencoder based approaches \cite{zhai2016deep,zhou2017anomaly,zong2018deep,zenati2018adversarially}, assume that outliers cannot be effectively reconstructed from the compressed low-dimensional projections. Clustering based methods \cite{xiong2011group,kim2012robust} aim at density estimation of data points and usually adopt a two-step strategy \cite{Varun2009survey} that performs dimensionality reduction firstly and then clustering. Different from previously mentioned categories, one-class classification based methods \cite{li2003improving,perdisci2006using,amer2013enhancing} make the effort to learn a discriminative boundary between the normal and abnormal instances.

\begin{figure*}%[!htbp]
  \centering
  \includegraphics[width=3in]{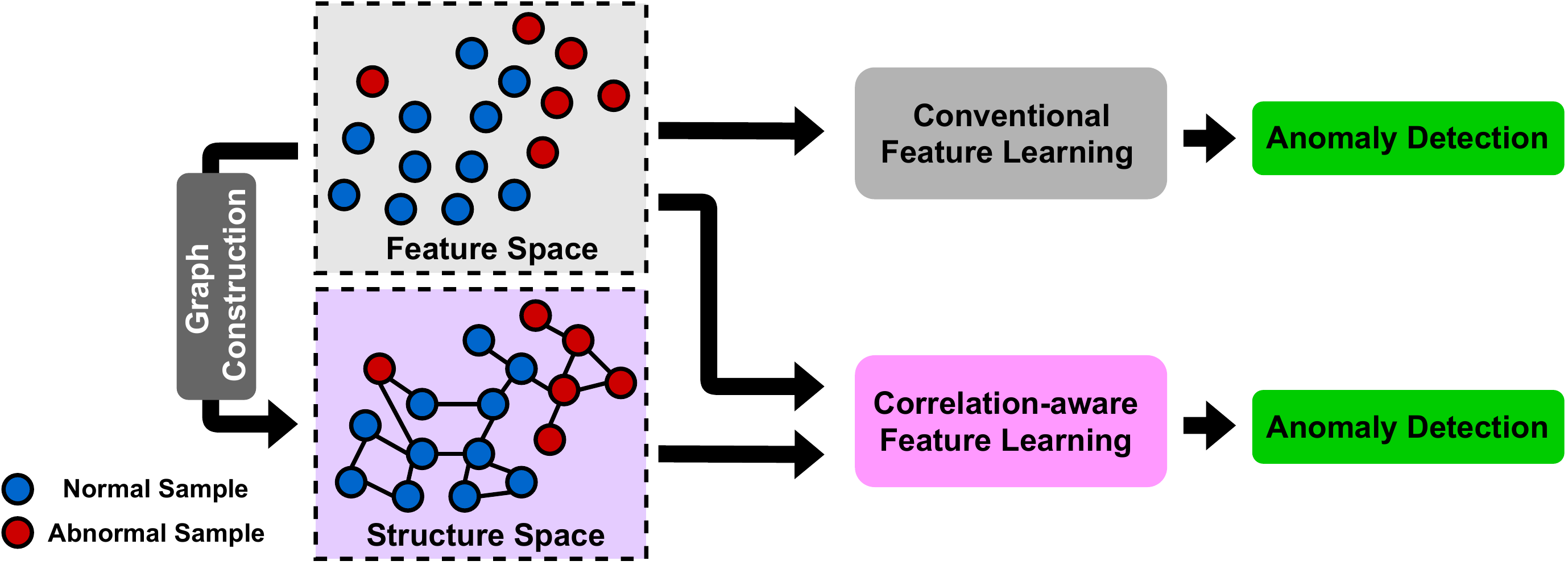}
 \caption{Correlation-aware feature learning for anomaly detection.}
\label{fig:motivation}
\end{figure*}

Although the above-mentioned methods had their fair share of success in anomaly detection, most of these methods neglect the complex correlation among data samples. As shown in Fig. \ref{fig:motivation}, the conventional methods attempt to conduct feature learning on the original observed feature space of data samples, while the correlation among similar samples is ignored, which can be exploited during feature learning by propagating more representative features from the neighbors to generate high-quality embedding for anomaly detection. However, modeling correlation among samples is far different from those conventional feature learning models, in which highly non-linear structure needs to be captured. Therefore, how to effectively incorporate both the original feature and relation structure of samples into an integrated feature learning framework for anomaly detection is still an open problem.

To alleviate the above-mentioned problems, in this paper, we propose a method of \textbf{C}orrelation aware unsupervised \textbf{A}nomaly detection via \textbf{D}eep \textbf{G}aussian \textbf{M}ixture \textbf{M}odel (\textbf{CADGMM}), which considers both the original feature and the complex correlation among data samples for feature learning. Specifically, the relations among data samples are correlated firstly in forms of a graph structure, in which, the node denotes the sample and the edge denotes the correlation between two samples from the feature space. Then, a dual-encoder that consists of a graph encoder and a feature encoder, is employed in CADGMM to encode both the feature and correlation of samples into the low-dimensional latent space jointly, followed by a decoder for data reconstruction. Finally, a separate estimation network as a Gaussian Mixture Model is utilized to estimate the density of the learned latent embedding. To verify the effectiveness of our algorithms, we conduct experiments on multiple real-world datasets. Our experimental results demonstrate that, by considering correlation among data samples, CADGMM significantly outperforms the state-of-the-art on unsupervised anomaly detection tasks.

\section{Notations and Problem Statement}
\label{sec:notations}

In this section, we formally define the frequently-used notations and the studied problem.

\begin{definition}
\textbf{Graph} is denoted as \textup{$\boldsymbol{\mathcal{G}}=\{\boldsymbol{\mathcal{V}}, \boldsymbol{\mathcal{E}}, \textbf{X} \}$} with \textup{$N$} nodes and \textup{$E$} edges, in which, \textup{$\mathcal{V}=\{v_i|i=1,2,...,N\}$} is a set of nodes, \textup{$\boldsymbol{\mathcal{E}}=\{e_i|i=1,2,...,E\}$} is a set of edges and \textup{$e_i=(v_{i_1}, v_{i_2})$} represents an edge between node \textup{$v_{i_1}$} and node \textup{$v_{i_2}$}.  \textup{$\textbf{X} \in {\mathbb{R}^{{N} \times {F}}}$} is an feature matrix with each row corresponding to a content feature of a node, where \textup{$F$} indicates the dimension of features. \textbf{Adjacency Matrix} of a graph is denoted as \textup{$\textbf{A} \in {\mathbb{R}^{{N} \times {N}}}$}, which can be used to represent the topologies of a graph. The scalar element \textup{$\textbf{A}_{i,j}=1$} if there exists an edge between node \textup{$v_i$} and node \textup{$v_j$}, otherwise, \textup{$\textbf{A}_{i,j}=0$}.
\end{definition}

\begin{problem}
\textbf{Anomaly detection}: Given a set of input samples $\boldsymbol{\mathcal{X}}=\{x_i|i=1,...,N\}$, each of which is associated with a $F$ dimension feature $\boldsymbol{\mathrm{X}}_i \in \mathbb{R}^{F}$, we aim to learn a score function $u(\boldsymbol{\mathrm{X}}_i): \mathbb{R}^{F} \mapsto \mathbb{R}$, to classify sample $x_i$ based on the threshold $\lambda$:
\begin{equation}
\label{eq:def}
\begin{split}
y_{i}=\left\{\begin{matrix}
1, & if \ u(\boldsymbol{\mathrm{X}}_i)\geq \lambda, \\ 
0, & otherwise.
\end{matrix}\right.
\end{split}
\end{equation}
where $y_{i}$ denotes the label of sample $x_i$, with 0 being the normal class and 1 the anomalous class.
\end{problem}

\begin{figure*}%[!htbp]
  \centering
\includegraphics[width=4.6in]{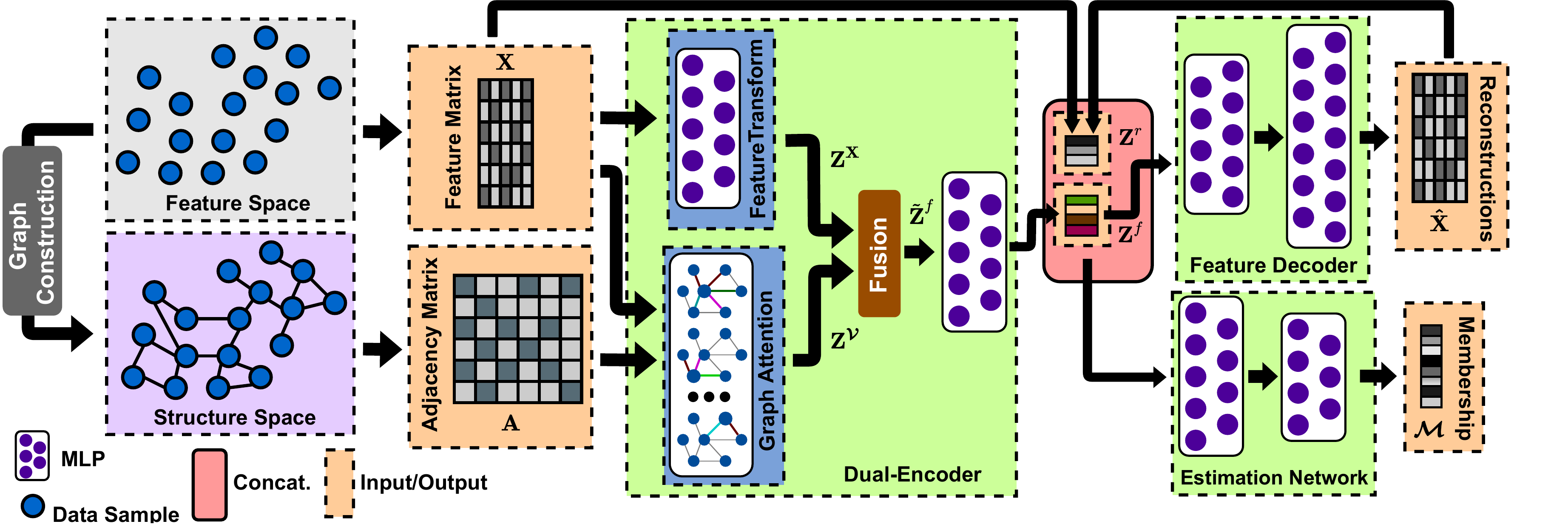}
 \caption{The framework of the proposed method.}
\label{fig:framework}
\end{figure*}

\section{Method}
\label{sec:method}
In this section, we introduce the proposed CADGMM in detail. CADGMM is an end-to-end joint representation learning framework for unsupervised anomaly detection. As shown in Fig. \ref{fig:framework}, CADGMM consists of three modules named dual-encoder, feature decoder, and estimation network, respectively. Specifically, the relations among data samples in the original feature space are correlated firstly in form of the graph structure. In the constructed graph, the node denotes the sample and the edge denotes the correlation between two samples in the feature space. Then, a dual-encoder that consists of a graph encoder and a feature encoder, is employed to encode both the feature and correlation information of samples into the low-dimensional latent space jointly, followed by a feature decoder for sample reconstruction. Finally, a separate estimation network is utilized to estimate the density of the learned latent embedding in the framework of Gaussian Mixture Model, and the anomalies can be detected by measuring the energy of the samples with respect to a given threshold.

\subsection{Graph Construction}
\label{subsec:graph_construction}
To explore the correlation among non-structure data samples for feature learning, we explicitly construct a graph structure to correlate the similar samples from the feature space. More specifically, given a set of input samples $\boldsymbol{\mathcal{X}}=\{x_i|i=1,...,N\}$, we employ \textbf{$K$-NN} algorithm on sample $x_i$ to determine its $K$ nearest neighbors $\boldsymbol{\mathcal{N}}_i=\{x_{i_k}|k=1,...,K\}$ in the feature space. Then, an undirected edge is assigned between $x_i$ and its neighbor $x_{i_k}$. Finally, an undirected graph $\boldsymbol{\mathcal{G}}=\{\boldsymbol{\mathcal{V}}, \boldsymbol{\mathcal{E}}, \boldsymbol{\mathrm{X}} \}$ is constructed, with $\boldsymbol{\mathcal{V}}=\{v_i=x_i |  i=1,...,N\}$ being the node set, $\boldsymbol{\mathcal{E}}=\{e_{i_k}=(v_{i}, v_{i_k}) | v_{i_k} \in \boldsymbol{\mathcal{N}}_i\}$ being the edge set, and $\boldsymbol{\mathrm{X}} \in \mathbb{R}^{N \times F}$ being the feature matrix of nodes. Based on the constructed graph, the feature affinities among samples are captured explicitly, which can be used during feature learning by performing message propagation mechanism on them.

\subsection{Dual-Encoder}
\label{subsec:dual_encoder}
In order to obtain sufficient representative high-level sample embedding, Dual-Encoder consists of a feature encoder and a graph encoder to encode the original feature of samples and the correlation among them respectively. 

To encode the original sample features $\boldsymbol{\mathrm{X}}$, feature encoder employs a $L_\mathrm{X}$ layers Multi-Layer Perceptron (MLP) to conduct a non-linear feature transform, which is as follows:
\begin{equation}
\begin{split}
\label{eq:feature_encoder}
&\textbf{Z}^{\boldsymbol{\mathrm{X}}(l_\mathrm{X})}=\sigma(\textbf{Z}^{\boldsymbol{\mathrm{X}}(l_\mathrm{X}-1)}\boldsymbol{\mathrm{W}}^{\boldsymbol{\mathrm{X}}(l_\mathrm{X}-1)}+{\boldsymbol{\textbf{b}}^{\boldsymbol{\mathrm{X}}(l_\mathrm{X}-1)}})
\end{split}
\end{equation}
where $\textbf{Z}^{\boldsymbol{\mathrm{X}}(l_\mathrm{X}-1)}$, $\textbf{Z}^{\boldsymbol{\mathrm{X}}(l_\mathrm{X})}$, $\textbf{W}^{\boldsymbol{\mathrm{X}}(l_\mathrm{X}-1)}$ and $\textbf{b}^{\boldsymbol{\mathrm{X}}(l_\mathrm{X}-1)}$ are the input, output, the trainable weight and bias matrix of ($l_\mathrm{X}$-$1$)-th layer respectively, $l_\mathrm{X}\in \{1,2,...,L_\mathrm{X}\}$, and $\textbf{Z}^{\boldsymbol{\mathrm{X}}(0)}=\textbf{X}$ is the initial input of the encoder. $\sigma(\bullet)$ denotes an activation function such as ReLU or Tanh. Finally, the final feature embedding $\textbf{Z}^{\boldsymbol{\mathrm{X}}}$=$\textbf{Z}^{\boldsymbol{\mathrm{X}}(L_\mathrm{X})}$ is obtained from the output of the last layer in MLP.

To encode the correlation among the samples, a graph attention layer \cite{velickovic2018graph} is employed to adaptively aggregate the representation from neighbor nodes, by performing a shared attentional mechanism on the nodes:
\begin{equation}
\begin{split}
\label{eq:init_node_transform}
& w_{i,j} = attn(\boldsymbol{\mathrm{X}}_{i}, \boldsymbol{\mathrm{X}}_{j})=\sigma(\textbf{a}^{\mathrm{T}}\cdot [\boldsymbol{\mathrm{W}}^{c} \boldsymbol{\mathrm{X}}_{i}||\boldsymbol{\mathrm{W}}^{c} \boldsymbol{\mathrm{X}}_{j}])
\end{split}
\end{equation}
where $w_{i,j}$ indicates the importance weight of node $\boldsymbol{v}_{i}$ to node $\boldsymbol{v}_{j}$, $attn(\bullet)$ denotes the neural network parametrized by weights $\textbf{a} \in \mathbb{R}^{D^{c}}$ and $\boldsymbol{\mathrm{W}}^{c} \in \mathbb{R}^{\frac{D^{c}}{2} \times F}$ that shared by all nodes and $D^{c}$ is the number of hidden neurons in $attn(\bullet)$, $||$ denotes the concatenate operation. Then, the final importance weight $\alpha_{i,j}$ is normalized through the softmax function:
\begin{equation}
\begin{split}
\label{eq:attn_norm}
&{\alpha_{i,j}}=\frac{\mathrm{exp}(w_{i,j})}{\sum_{k \in \boldsymbol{\mathcal{N}_{i}}}\mathrm{exp}(w_{i,k})}
\end{split}
\end{equation}
where $\boldsymbol{\mathcal{N}_{i}}$ denotes the neighbors of node $\boldsymbol{v}_i$, which is provided by adjacency matrix \textup{$\boldsymbol{\mathrm{A}}$}, and the final node embedding $\boldsymbol{{\mathrm{Z}}^{\mathcal{V}}}=\{\boldsymbol{{\mathrm{Z}}_{i}^{\mathcal{V}}}\}$ can be obtained by the weighted sum based on the learned importance weights as follows:
\begin{equation}
\begin{split}
\label{eq:final_node_embedding}
&\boldsymbol{{\mathrm{Z}}_{i}^{\mathcal{V}}}=\sum_{k \in \boldsymbol{\mathcal{N}_{i}}}\alpha_{i,k}\cdot \boldsymbol{\mathrm{X}}_{k}
\end{split}
\end{equation}

Given the learned embedding $\textbf{Z}^{\boldsymbol{\mathrm{X}}}$ and $\boldsymbol{{\mathrm{Z}}^{\mathcal{V}}}$, a fusion module is designed to fuse the embeddings from heterogeneous data source into a shared latent space, followed by a fully connected layer to obtain the final sample embedding $\textbf{Z}^{f} \in \mathbb{R}^{N \times D}$:
\begin{equation}
\begin{split}
\label{eq:fuse}
&\Tilde{\textbf{Z}}^{f}=\text{Fusion}(\textbf{Z}^{\boldsymbol{\mathrm{X}}},\boldsymbol{{\mathrm{Z}}^{\mathcal{V}}})=\textbf{Z}^{\boldsymbol{\mathrm{X}}}\oplus \boldsymbol{{\mathrm{Z}}^{\mathcal{V}}}
\end{split}
\end{equation}
\begin{equation}
\begin{split}
\label{eq:final_encoder}
&\textbf{Z}^{f}=\Tilde{\textbf{Z}}^{f}\boldsymbol{\mathrm{W}}+{\boldsymbol{\textbf{b}}}
\end{split}
\end{equation}
where $\textbf{W}$ and $\textbf{b}$ are the trainable weight and bias matrix, and $\oplus$ indicates the element-wise plus operator of two matrices.

\subsection{Feature Decoder}
\label{subsec:feature_decoder}
Feature decoder aims at reconstructing the sample features from the latent embedding $\textbf{Z}^{f}$:
\begin{equation}
\begin{split}
\label{eq:feature_encoder}
&\textbf{Z}^{\boldsymbol{\hat{\mathrm{X}}}(l_{\hat{\mathrm{X}}})}=\sigma(\textbf{Z}^{\boldsymbol{\hat{\mathrm{X}}}(l_{\hat{\mathrm{X}}}-1)}\textbf{W}^{\boldsymbol{\hat{\mathrm{X}}}(l_{\hat{\mathrm{X}}}-1)}+\textbf{b}^{\boldsymbol{\hat{\mathrm{X}}}(l_{\hat{\mathrm{X}}}-1)})
\end{split}
\end{equation}
where $\textbf{Z}^{\boldsymbol{\hat{\mathrm{X}}}(l_{\hat{\mathrm{X}}}-1)}$, $\textbf{Z}^{\boldsymbol{\hat{\mathrm{X}}}(l_{\hat{\mathrm{X}}})}$, $\textbf{W}^{\boldsymbol{\hat{\mathrm{X}}}(l_{\hat{\mathrm{X}}}-1)}$ and $\textbf{b}^{\boldsymbol{\hat{\mathrm{X}}}(l_{\hat{\mathrm{X}}}-1)}$ are the input, output, the trainable weight and bias matrix of ($l_{\hat{\mathrm{X}}}$-$1$)-th layer of decoder respectively, $l_{\hat{\mathrm{X}}}\in\{1,2,...,L_{\hat{\mathrm{X}}}\}$, and $\textbf{Z}^{\boldsymbol{\hat{\mathrm{X}}}(0)}=\textbf{Z}^{f}$ is the initial input of the decoder. 
Finally, the reconstruction $\boldsymbol{\hat{\mathrm{X}}}$ is obtained from the last layer of decoder:
\begin{equation}
\begin{split}
\label{eq:feature_encoder}
\boldsymbol{\hat{\mathrm{X}}}=\textbf{Z}^{\boldsymbol{\hat{\mathrm{X}}}(L_{\hat{\mathrm{X}}})}
\end{split}
\end{equation}

\subsection{Estimate Network}
\label{subsec:estimate_network}
To estimate the density of the input samples, a Gaussian Mixture Model is leveraged in CADGMM over the learned latent embedding. Inspired by DAGMM \cite{zong2018deep}, a sub-network consists of several fully connected layers is utilized, which takes the reconstruction error preserved low-dimentional embedding as input, to estimate the mixture membership for each sample. The reconstruction error preserved low-dimentional embedding $\textbf{Z}$ is obtained as follows:
\begin{equation}
\begin{split}
\label{eq:final_embedding}
&\textbf{Z}=[\textbf{Z}^{f}||\textbf{Z}^{r}], \ \textbf{Z}^{r}=\text{Dist}(\textbf{X}, \boldsymbol{\hat{\mathrm{X}}})
\end{split}
\end{equation}
where $\textbf{Z}^{r}$ is the reconstruction error embedding and $\text{Dist}(\bullet)$ denotes the distance metric such as Euclidean distance or cosine distance. Given the final embedding $\textbf{Z}$ as input, estimate network conducts membership prediction as follows:
\begin{equation}
\begin{split}
\label{eq:estimate_network}
&\textbf{Z}^{\boldsymbol{\mathcal{M}}(l_\mathcal{M})}=\sigma(\textbf{Z}^{\boldsymbol{\mathcal{M}}(l_\mathcal{M}-1)}\textbf{W}^{\boldsymbol{\mathcal{M}}(l_\mathcal{M}-1)}+\textbf{b}^{\boldsymbol{\mathcal{M}}(l_\mathcal{M}-1)})
\end{split}
\end{equation}
where $\textbf{Z}^{\boldsymbol{\mathcal{M}}(l_\mathcal{M}-1)}$, $\textbf{Z}^{\boldsymbol{\mathcal{M}}(l_\mathcal{M})}$, $\textbf{W}^{\boldsymbol{\mathcal{M}}(l_\mathcal{M}-1)}$ and $\textbf{b}^{\boldsymbol{\mathcal{M}}(l_\mathcal{M}-1)}$ are the input, output, the trainable weight and bias matrix of ($l_\mathcal{M}$-$1$)-th layer of estimate network respectively, $l_\mathcal{M}\in\{1,2,...,L_\mathcal{M}\}$, $\textbf{Z}^{\boldsymbol{\mathcal{M}}(0)}=\textbf{Z}$, and the mixture-component membership $\boldsymbol{\mathcal{M}}$ is calculated by:
\begin{equation}
\begin{split}
\label{eq:membership}
\boldsymbol{\mathcal{M}}=\text{Softmax}(\textbf{Z}^{\boldsymbol{\mathcal{M}}(L_\mathcal{M})})
\end{split}
\end{equation}
where $\boldsymbol{\mathcal{M}} \in  \mathbb{R}^{N\times M}$ is the predicted membership of $M$ mixture components for $N$ samples. With the predicted sample membership, the parameters of GMM can be calculated to facilitate the evaluation of the energy/likelihood of input samples, which is as follows:
\begin{equation}
\begin{split}
\label{eq:means_covariance}
\boldsymbol{\mu_{m}}=\frac{\sum_{i=1}^{N}\boldsymbol{\mathcal{M}}_{i,m}\textbf{Z}_{i}}{\sum_{i=1}^{N}\boldsymbol{\mathcal{M}}_{i,m}},\ \boldsymbol{\Sigma}_{m}=\frac{\sum_{i=1}^{N}\boldsymbol{\mathcal{M}}_{i,m}(\textbf{Z}_{i}-\boldsymbol{\mu_{m}})(\textbf{Z}_{i}-\boldsymbol{\mu_{m}})^{\text{T}}}{\sum_{i=1}^{N}\boldsymbol{\mathcal{M}}_{i,m}}
\end{split}
\end{equation}
where $\boldsymbol{\mu_{m}}$ and $\boldsymbol{\Sigma}_{m}$ are the means and covariance of the $m$-th component distribution respectively, and the energy of samples is as follows:
\begin{equation}
\begin{split}
\label{eq:energy}
\text{E}_{\textbf{Z}}=-\text{log}\left ( \sum_{m=1}^{M}\sum_{i=1}^{N}\frac{\boldsymbol{\mathcal{M}}_{i,m}}{N} \frac{\text{exp}(-\frac{1}{2}(\textbf{Z}-\boldsymbol{\mu_{m}})^{\text{T}}\boldsymbol{\Sigma}_{m}^{-1}(\textbf{Z}-\boldsymbol{\mu_{m}}))}{|2\pi\boldsymbol{\Sigma}_{m}|^{\frac{1}{2}}}  \right )
\end{split}
\end{equation}
where $|\bullet|$ indicates the determinant of a matrix.

\subsection{Loss Function and Anomaly Score}
\label{sec:loss_function}
The training objective of CADGMM is defined as follows:
\begin{equation}
\label{eq:loss}
\begin{split}
\mathcal{L}=||\textbf{X}-\boldsymbol{\hat{\mathrm{X}}}||_{2}^{2}+\lambda_{1}E_{\textbf{Z}}+
\lambda_{2}\sum_{m=1}^{M}\sum_{i=1}^{N}\frac{1}{(\boldsymbol{\Sigma}_{m})_{ii}}+\lambda_{3}||\textbf{Z}||_{2}^{2}
\end{split}
\end{equation}
where the first term is reconstruction error used for feature reconstruction, the second is sample energy, which aims to maximize the likelihood to observed samples, the third is covariance penalization, used for solving singularity problem as in GMM \cite{zong2018deep} by penalizing small values on the diagonal entries of covariance matrix, and the last is embedding penalization, which serves as a regularizer to impose the magnitude of normal samples as small as possible in the latent space, to deviate the normal samples from the abnormal ones. $\lambda_{1}$, $\lambda_{2}$, and $\lambda_{3}$ are three parameters which control the trade off between different terms.

The anomaly score is the sample energy $E_{\textbf{Z}}$, and based on the measured anomaly scores, the threshold $\lambda$ in Eq. \ref{eq:def} can be determined according to the distribution of scores, e.g. the samples of top-k scores are classified as anomalous samples.

\section{Experiments}
\label{sec:experiments}
In this section, we  will describe the experimental details including datasets, baseline methods, and parameter settings, respectively.

\begin{table}[t!]
\caption{Statistics of the public benchmark datasets.}
\label{tab:datasets}
\centering
\begin{tabular}{|c|c|c|c|}
\hline
Database	& \# Dimensions	&\# Instances &\ Anomaly ratio \\
\hline
\hline
KDD99    &120  &494,021	 &0.2 	\\
Arrhythmia	&274  &452 	&0.15 	\\
Satellite   &36	&6,435 &0.32 \\
\hline	
\end{tabular}
\end{table}

\subsection{Dataset}
\label{sebsec:dataset}

Three benchmark datasets are used in this paper to evaluate the proposed method, including KDD99, Arrhythmia, and Satellite. The statistics of datasets are shown in Table \ref{tab:datasets}.

\begin{itemize}
\item \textbf{KDD99} The KDD99 10 percent dataset \cite{bache2013uci} contains 494021 samples with 41 dimensional features, where 34 of them are continuous and 7 are categorical. One-hot representation is used to encode the categorical features, resulting in a 120-dimensional feature for each sample. 
\item \textbf{Arrhythmia} The Arrhythmia dataset \cite{bache2013uci} contains 452 samples with 274 dimensional features. We combine the smallest classes including 3, 4, 5, 7, 8, 9, 14, 15 to form the outlier class and the rest of the classes are inliers class.
\item \textbf{Satellite} The Satellite dataset \cite{bache2013uci} has 6435 samples with 36 dimensional features. The smallest three classes including 2,4,5 are combined to form the outliers and the rest are inliers classes.

\end{itemize}

\subsection{Baseline Methods}
\label{sebsec:baselines}
\begin{itemize}
\item \textbf{One Class Support Vector Machines (OC-SVM)} \cite{chen2001one} is a classic kernel method for anomaly detection, which learns a decision boundary between the inliers and outliers.

\item \textbf{Isolation Forests (IF)} \cite{liu2008isolation} conducts anomaly detection by building trees using randomly selected split values across sample features, and defining the anomaly score as the average path length from a specific sample to the root.

\item \textbf{Deep Structured Energy Based Models (DSEBM)} \cite{zhai2016deep} is a deep energy-based model, which aims to accumulate the energy across the layers. DSEBM-r and DSEBM-e are utilized in \cite{zhai2016deep} by taking the energy and reconstruction error as the anomaly score respectively.

\item \textbf{Deep Autoencoding Gaussian Mixture Model (DAGMM)} \cite{zong2018deep} is an autoencoder based method for anomaly detection, which consists of a compression network for dimension reduction, and an estimate network to perform density estimation under the Gaussian Mixture Model.

\item \textbf{AnoGAN} \cite{schlegl2017unsupervised} is an anomaly detection algorithm based on GAN, which trains a DCGAN \cite{radford2015unsupervised} to recover the representation of each data sample in the latent space during prediction. 
\item \textbf{ALAD} \cite{zenati2018adversarially} is based on bi-directional GANs for anomaly detection by deriving adversarially learned features and uses reconstruction errors based on the learned features to determine if a data sample is anomalous.

\end{itemize}

\subsection{Parameter Settings}
\label{subsec:para_settings}
The parameter settings in the experiment for different datasets are as follows:
\begin{itemize}
\item \textbf{KDD99} For KDD99, CADGMM is trained with 300 iterations and $N$=1024 for graph construction with $K$=15, which is the batch size for training. $M$=4, $\lambda_1$=0.1, $\lambda_2$=0.005, $\lambda_3$=10.
\item \textbf{Arrhythmia} For Arrhythmia, CADGMM is trained with 20000 iterations and $N$=128 for graph construction with $K$=5, which is the batch size for training, $M$=2, $\lambda_1$=0.1, $\lambda_2$=0.005, $\lambda_3$=0.001.
\item \textbf{Satellite} For Satellite, CADGMM is trained with 3000 iterations and $N$=512 for graph construction with $K$=13, $M$=4, $\lambda_1$=0.1, $\lambda_2$=0.005, $\lambda_3$=0.005.
\end{itemize}
The architecture details of CADGMM on different datasets are shown in Table \ref{tab:architecture}, in which, $\text{FC}(D_{in}, D_{out})$ means a fully connected layer with $D_{in}$ input neurons and $D_{out}$ output neurons. Similarly, $\text{GAT}(D_{in}, D_{out})$ means a graph attention layer with $D_{in}$-dimensional input and $D_{out}$-dimensional output. The activation function $\sigma(\bullet)$ for all datasets is set as Tanh.
For the baseline methods, we set the parameters by grid search. We independently run each experiment 10 times and the mean values are reported as the final results.

\begin{table*}%[htp!]
\caption{Architecture details of CADGMM for different datasets.}
\label{tab:architecture}
\centering
\begin{tabular}{V{2}cV{2}c|c|c|c|cV{2}}
\hlineB{2}

\multirow{2}{*}{Dataset}
&\multicolumn{3}{c|}{Dual-Enc.} &\multirow{2}{*}{Feature Dec.} &\multirow{2}{*}{Estimate Net.} \\\cline{2-4}

&Feature Trans.   &Graph Attn. &MLP & & \\

\hlineB{2}

\multirow{3}{*}{KDD99} &FC(120,64) &GAT(120,32) &FC(32, 8) &FC(8,32) &FC(10,20) \\

&FC(64,32) & & &FC(32,64) &FC(20,8) \\

& & & &FC(64,120) &FC(8,4) \\

\hlineB{2}

\multirow{2}{*}{Arrhythmia} &FC(274,32) &GAT(274,32) &FC(32, 2) &FC(2,10) &FC(4,10) \\

& & & &FC(10,274) &FC(10,2) \\

\hlineB{2}

\multirow{2}{*}{Satellite} &FC(36,16) &GAT(36,16) &FC(16, 2) &FC(2,16) &FC(4,10) \\

& & & &FC(16,36) &FC(10,4) \\

\hlineB{2}

\end{tabular}
\end{table*}

\begin{table*}%[htp!]
\caption{Anomaly Detection Performance on KDD99, Arrhythmia, and Satellite datasets. Better results are marked in \textbf{bold}.}
\label{tab:anomaly_detection}
\centering
\begin{tabular}{V{2}cV{2}c|c|c|c|c|c|c|c|cV{2}}
\hlineB{2}

\multirow{2}{*}{Method}
&\multicolumn{3}{c|}{KDD99} &\multicolumn{3}{c|}{Arrhythmia} &\multicolumn{3}{cV{2}}{Satellite} \\\cline{2-10}

&Precision   &Recall &F1 &Precision   &Recall &F1 &Precision   &Recall &F1   \\
\hlineB{2}
OC-SVM \cite{chen2001one} &74.57 &85.23 &79.54 &53.97 &40.82 &45.81 &52.42 &59.99 &61.07 \\
IF \cite{liu2008isolation} &92.16 &93.73 &92.94 &51.47 &54.69 &53.03 &60.81 &\textbf{94.89} &75.40  \\
DSEBM-r \cite{zhai2016deep} &85.21 &64.72 &73.28 &15.15 &15.13 &15.10 &67.84 &68.61 &68.22 \\
DSEBM-e \cite{zhai2016deep} &86.19 &64.66 &73.99 &46.67 &45.65 &46.01 &67.79 &68.56 &68.18 \\
DAGMM \cite{zong2018deep} &92.97 &94.42 &93.69 &49.09 &50.78 &49.83 &80.77 &81.6 &81.19 \\
AnoGAN \cite{schlegl2017unsupervised} &87.86 &82.97 &88.65 &41.18 &43.75 &42.42 &71.19 &72.03 &71.59 \\
ALAD \cite{zenati2018adversarially} &94.27 &95.77 &95.01 &50 &53.13 &51.52 &79.41 &80.32 &79.85 \\
\hline
\textbf{CADGMM} &\textbf{96.01} &\textbf{97.53} &\textbf{96.71} &\textbf{56.41} &\textbf{57.89} &\textbf{57.14} &\textbf{81.99} &82.75 &\textbf{82.37} \\

\hlineB{2}
\end{tabular}
\end{table*}

\section{Results and Analysis}
\label{sec:results_analysis}
In this section, we will demonstrate the effectiveness of the proposed method by presenting results of our model on anomaly detection task, and provide a comparison with the state-of-the-art methods.

\subsection{Anomaly Detection}
\label{subsec:anomaly_detection_task}

As in previous literatures \cite{zhai2016deep,zong2018deep,zenati2018adversarially}, in this paper, \textbf{Precision}, \textbf{Recall} and \textbf{F1} score are employed as the evaluation metrics. Generally, we expect the values of these evaluation metrics as big as possible. The sample with high energy is classified as abnormal and the threshold is determined based on the ratio of anomalies in the dataset. Following the settings in \cite{zhai2016deep,zong2018deep}, the training and test sets are split by 1:1 and only normal samples are used for training the model.

The experimental results shown in Table \ref{tab:anomaly_detection} demonstrate that the proposed CADGMM significantly outperforms all baselines in various datasets. The performance of CADGMM is much higher than traditional anomaly detection methods such as OC-SVM and IF, because of the limited capability of feature learning or the curse of dimensionality. Moreover, CADGMM also significantly outperforms all other deep learning based methods such as DSEBM, DAGMM, AnoGAN, and ALAD, which demonstrates that additional correlation among data samples facilitates the feature learning for anomaly detection. For small datasets such as Arrhythmia, we can find that traditional methods such as IF are competitive compared with conventional deep learning based method such as DSEBM, DAGMM, AnoGAN, and ALAD, which might because that the lack of sufficient training data could have resulted in poorer performance of the data hungry deep learning based methods, while CADGMM is capable of leveraging more data power given the limited data source, by considering the correlation among data samples.

\begin{table*}%[htp!]
\caption{Anomaly Detection Performance on KDD99 with different ratios of anomalies during training.}
\label{tab:noise}
\centering
\begin{tabular}{V{2}cV{2}c|c|c|c|c|c|c|c|cV{2}}
\hlineB{2}
\multirow{2}{*}{Radio}
&\multicolumn{3}{c|}{CADGMM} &\multicolumn{3}{c|}{DAGMM} &\multicolumn{3}{cV{2}}{OC-SVM} \\\cline{2-10}

&Precision   &Recall &F1 &Precision   &Recall &F1 &Precision   &Recall &F1 \\
\hlineB{2}
1\% &95.53 &97.04 &96.28 &92.01 &93.37 &92.68 &71.29 &67.85 &69.53\\
2\% &95.32 &96.82 &96.06 &91.86 &93.40 &92.62 &66.68 &52.07 &58.47\\
3\% &94.83 &96.33 &95.58 &91.32 &92.72 &92.01 &63.93 &44.70 &52.61\\
4\% &94.62 &96.12 &95.36 &88.37 &89.89 &89.12 &59.91 &37.19 &45.89\\
5\% &94.35 &96.04 &95.3 &85.04 &86.43 &85.73 &11.55 &33.69 &17.20\\

\hline
\hlineB{2}
\end{tabular}
\end{table*}

\subsection{Impact of noise data}
\label{subsec:impact_noise}
In this section, we study the impact of noise data for the training of CADGMM. To be specific, 50\% of randomly split data samples are used for testing, while the rest 50\% combined with 1\% to 5\%  anomalies are used for training.

As shown in Table \ref{tab:noise}, with the increase of noise data, the performance of all baselines degrade significantly, especially for OC-SVM, which tends to be more sensitive to noise data because of its poor ability of feature learning on high-dimensional data. However, CADGMM performs stable with different ratios of noise and achieves state-of-the-art even 5\% anomalies are injected into the training data, which demonstrates the robustness of the proposed method.

\begin{figure*}%[!htbp]
  \centering
\includegraphics[width=4.2in]{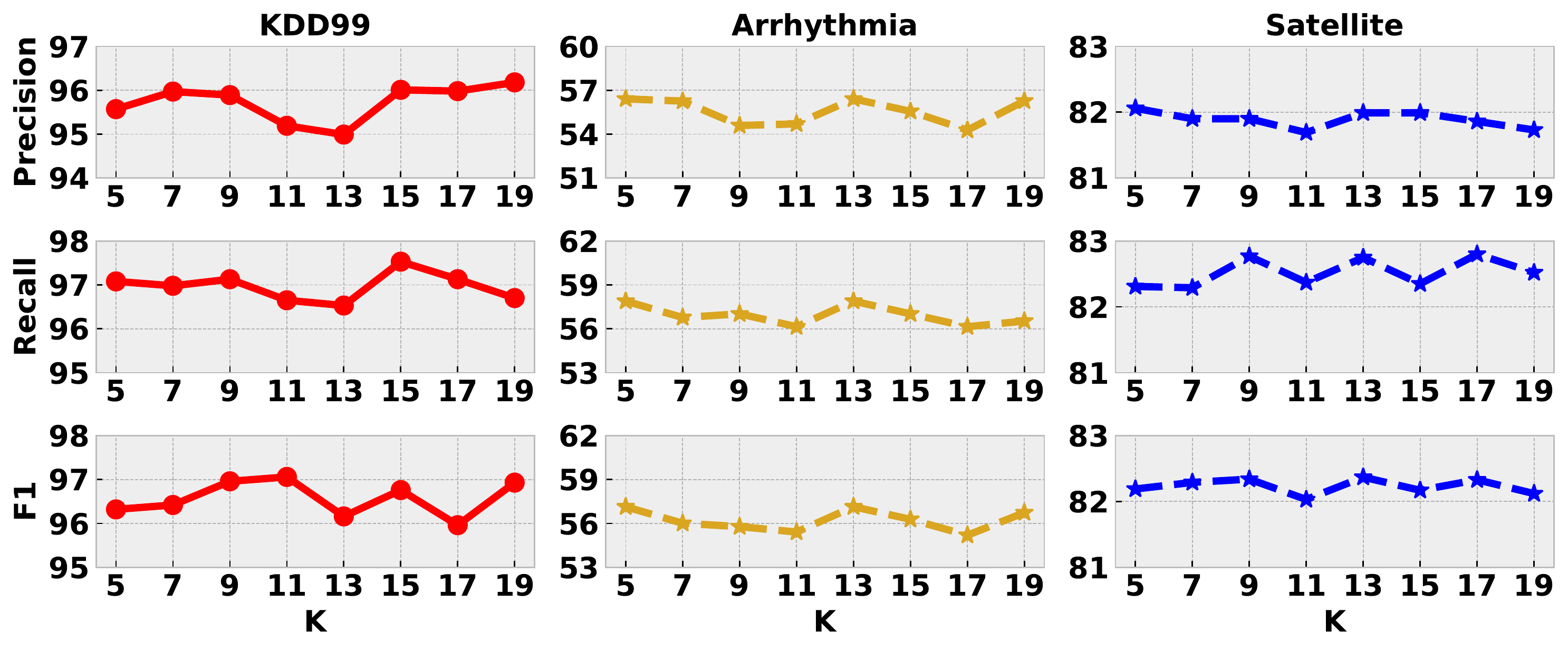}
 \caption{Impact of different $K$ values of K-NN algorithms in graph construction.}
\label{fig:impact_knn}
\end{figure*}

\begin{figure}%[!htbp]
  \centering
  \subfloat[DAGMM]{\includegraphics[width=1.2in]{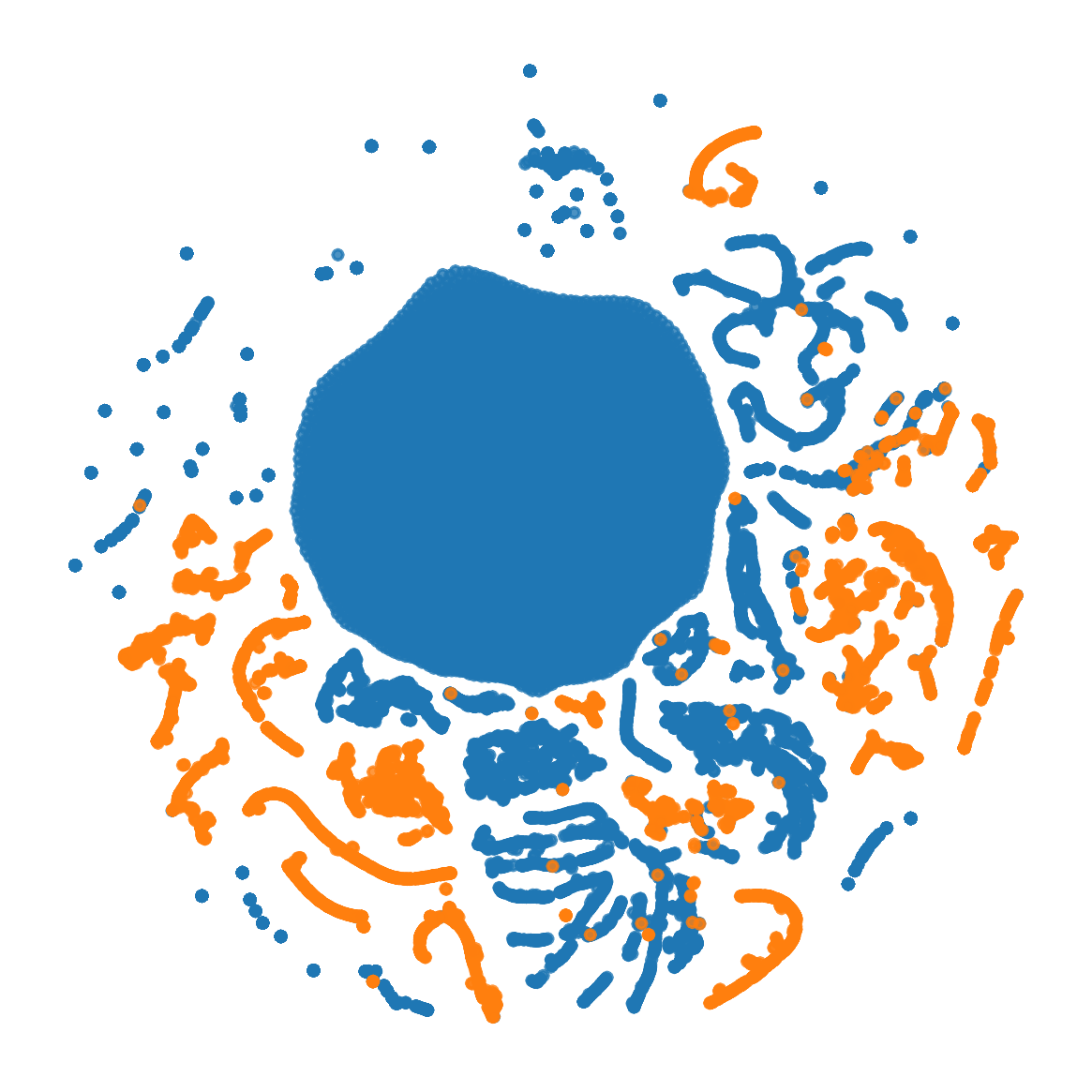}
\label{fig:dagmm_vis_40000}}
\hfil
  \subfloat[CADGMM]{\includegraphics[width=1.2in]{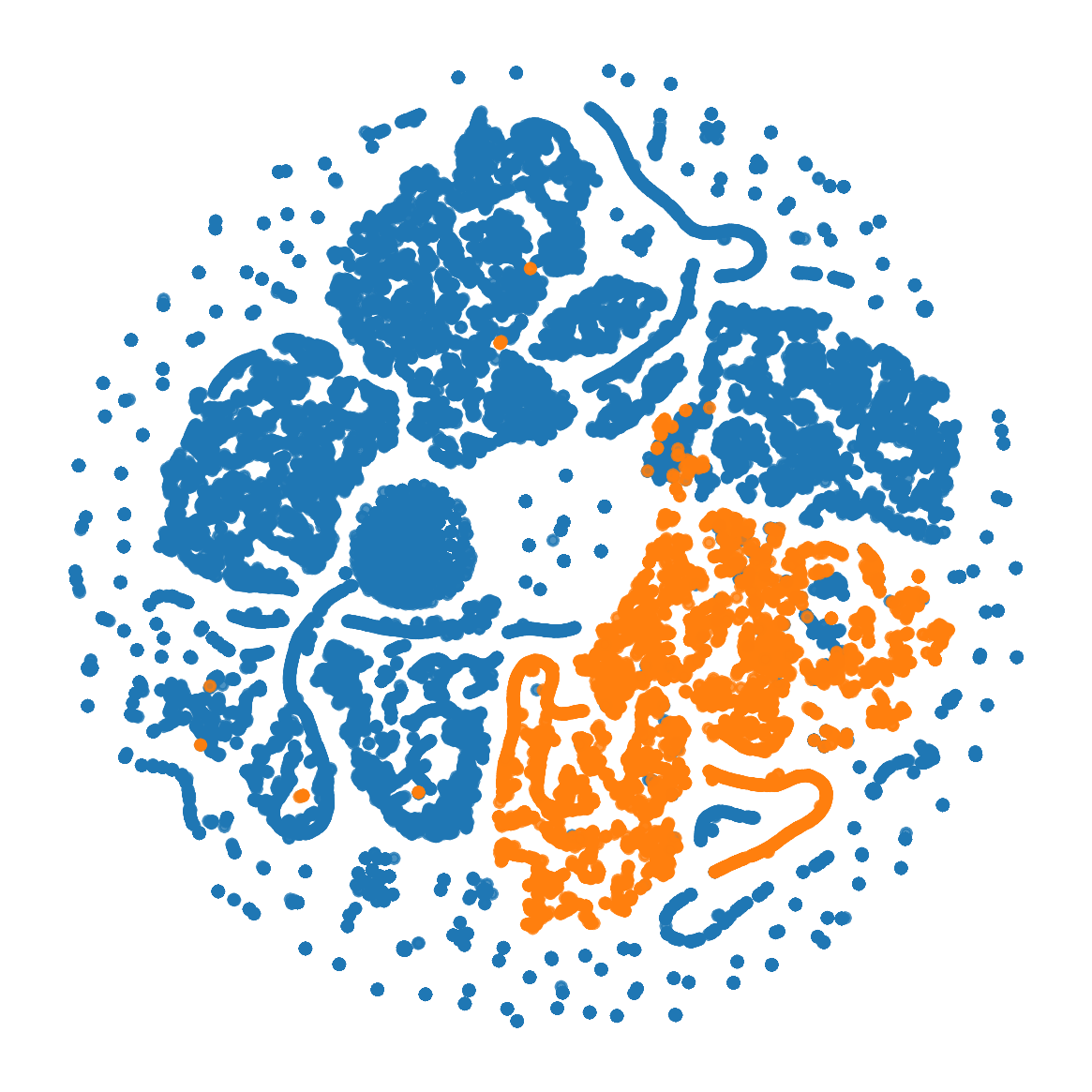}
\label{fig:cadgmm_vis_40000}}
  \caption{Embedding visualization (Blue indicates the normal samples and orange the anomalies).}
\label{fig:visualization}
\end{figure}

\subsection{Impact of $K$ values}
\label{subsec:impact_KNN}
In this section, we evaluate the impact of different $K$ values during the graph construction on CADGMM.

More specifically, we conduct experiments on all three datasets by varying the number of $K$ from 5 to 19, and the experimental results are illustrated in Fig. \ref{fig:impact_knn}. During training, the batch sizes are set as 1024, 128, and 512 for KDD99, Arrhythmia, and Satellite, respectively, the experimental results show that the changing of $K$ value causes only a little fluctuation of performance on all datasets with different settings, which demonstrates that CADGMM is less sensitive to the $K$ value and easy to use.

\subsection{Embedding Visualization}
\label{subsec:visualization}

In order to explore the quality of the learned embedding, we make a comparison of the visualization of sample representation for different methods in Fig. \ref{fig:visualization}. Specifically, we take the low-dimensional embeddings of samples learned by DAGMM and CADGMM, as the inputs to the t-SNE tool \cite{maaten2008visualizing}. Here, we randomly choose 40000 data samples from the test set of KDD99 for visualization, and then we generate visualizations of the sample embedding on a two-dimensional space, in which blue colors correspond to the normal class while orange the abnormal class. We can find that CADGMM achieves more compact and separated clusters compared with DAGMM. The results can also explain why our approach achieves better performance on anomaly detection task.

\section{Conclusion}
\label{sec:conclusion}
In this paper, we study the problem of correlation aware unsupervised anomaly detection, which considers the correlation among data samples from the feature space. To cope with this problem, we propose a method named CADGMM to model the complex correlation among data points to generate high-quality low-dimensional embeddings for anomaly detection. Extensive experiments on real-world datasets demonstrate the effectiveness of the proposed method.

\section*{Acknowledgement}

This work was supported in part by National Natural Science Foundation of China (No. 61172168, 61972187).

\bibliographystyle{splncs04}
\bibliography{refs}

\begin{thebibliography}{10}
\providecommand{\url}[1]{\texttt{#1}}
\providecommand{\urlprefix}{URL }
\providecommand{\doi}[1]{https://doi.org/#1}

\bibitem{amer2013enhancing}
Amer, M., Goldstein, M., Abdennadher, S.: Enhancing one-class support vector
  machines for unsupervised anomaly detection. In: SIGKDD. pp. 8--15 (2013)

\bibitem{bache2013uci}
Bache, K., Lichman, M.: Uci machine learning repository, 2013. URL
  http://archive. ics. uci. edu/ml  \textbf{5} (2013)

\bibitem{Varun2009survey}
Chandola, V., Banerjee, A., Kumar, V.: Anomaly detection: A survey. ACM
  Computing Surveys  \textbf{41}(3) (7 2009). \doi{10.1145/1541880.1541882}

\bibitem{chen2001one}
Chen, Y., Zhou, X.S., Huang, T.S.: One-class svm for learning in image
  retrieval. In: ICIP. pp. 34--37 (2001)

\bibitem{jolliffe2003principal}
Jolliffe, I.: Principal component analysis. Technometrics  \textbf{45}(3), ~276
  (2003)

\bibitem{kim2012robust}
Kim, J., Scott, C.D.: Robust kernel density estimation. Journal of Machine
  Learning Research  \textbf{13}(Sep),  2529--2565 (2012)

\bibitem{li2003improving}
Li, K.L., Huang, H.K., Tian, S.F., Xu, W.: Improving one-class svm for anomaly
  detection. In: Proceedings of the 2003 International Conference on Machine
  Learning and Cybernetics. vol.~5, pp. 3077--3081 (2003)

\bibitem{liu2008isolation}
Liu, F.T., Ting, K.M., Zhou, Z.H.: Isolation forest. In: ICDM. pp. 413--422
  (2008)

\bibitem{maaten2008visualizing}
Maaten, L.v.d., Hinton, G.: Visualizing data using t-sne. Journal of machine
  learning research  \textbf{9}(Nov),  2579--2605 (2008)

\bibitem{pascoal2012robust}
Pascoal, C., De~Oliveira, M.R., Valadas, R., Filzmoser, P., Salvador, P.,
  Pacheco, A.: Robust feature selection and robust pca for internet traffic
  anomaly detection. In: IEEE INFOCOM. pp. 1755--1763 (2012)

\bibitem{perdisci2006using}
Perdisci, R., Gu, G., Lee, W., et~al.: Using an ensemble of one-class svm
  classifiers to harden payload-based anomaly detection systems. In: ICDM.
  vol.~6, pp. 488--498 (2006)

\bibitem{radford2015unsupervised}
Radford, A., Metz, L., Chintala, S.: Unsupervised representation learning with
  deep convolutional generative adversarial networks. ICLR  (2016)

\bibitem{schlegl2017unsupervised}
Schlegl, T., Seeb{\"o}ck, P., Waldstein, S.M., Schmidt-Erfurth, U., Langs, G.:
  Unsupervised anomaly detection with generative adversarial networks to guide
  marker discovery. In: IPMI. pp. 146--157 (2017)

\bibitem{sultani2018real}
Sultani, W., Chen, C., Shah, M.: Real-world anomaly detection in surveillance
  videos. In: CVPR. pp. 6479--6488 (2018)

\bibitem{tan2011fast}
Tan, S.C., Ting, K.M., Liu, T.F.: Fast anomaly detection for streaming data.
  In: IJCAI (2011)

\bibitem{velickovic2018graph}
Veli{\v{c}}kovi{\'{c}}, P., Cucurull, G., Casanova, A., Romero, A., Li{\`{o}},
  P., Bengio, Y.: Graph attention networks (2018)

\bibitem{xiong2011group}
Xiong, L., P{\'o}czos, B., Schneider, J.G.: Group anomaly detection using
  flexible genre models. In: NIPS. pp. 1071--1079 (2011)

\bibitem{xu2010robust}
Xu, H., Caramanis, C., Sanghavi, S.: Robust pca via outlier pursuit. In: NIPS.
  pp. 2496--2504 (2010)

\bibitem{xu2018cxnet}
Xu, S., Wu, H., Bie, R.: Cxnet-m1: Anomaly detection on chest x-rays with
  image-based deep learning. IEEE Access  \textbf{7},  4466--4477 (2018)

\bibitem{zenati2018adversarially}
Zenati, H., Romain, M., Foo, C.S., Lecouat, B., Chandrasekhar, V.:
  Adversarially learned anomaly detection. In: ICDM. pp. 727--736 (2018)

\bibitem{zhai2016deep}
Zhai, S., Cheng, Y., Lu, W., Zhang, Z.: Deep structured energy based models for
  anomaly detection. In: ICML. pp. 1100--1109 (2016)

\bibitem{zhou2017anomaly}
Zhou, C., Paffenroth, R.C.: Anomaly detection with robust deep autoencoders.
  In: SIGKDD. pp. 665--674 (2017)

\bibitem{zong2018deep}
Zong, B., Song, Q., Min, M.R., Cheng, W., Lumezanu, C., Cho, D., Chen, H.: Deep
  autoencoding gaussian mixture model for unsupervised anomaly detection. In:
  ICLR (2018)

\end{thebibliography}

\end{document}